\documentclass[11pt]{article}

\usepackage[preprint]{acl}

\author{Brady Steele \\
    Georgia Institute of Technology \\
    \texttt{bsteele45@gatech.edu}}

\usepackage{times}
\usepackage{latexsym}

\usepackage[T1]{fontenc}
\usepackage[utf8]{inputenc}
\usepackage{microtype}
\usepackage{inconsolata}

\usepackage{amsmath,amssymb}
\usepackage{booktabs}
\usepackage{graphicx}
\graphicspath{{./}}
\usepackage{multirow}
\usepackage{xcolor}
\usepackage{float}

\newcommand{\reals}{\mathbb{R}}
\newcommand{\loraA}{\mathbf{A}}
\newcommand{\loraB}{\mathbf{B}}
\newcommand{\deltaW}{\Delta\mathbf{W}}
\newcommand{\entropy}{\mathcal{H}}
\newcommand{\aulc}{\text{AULC}}

\title{Annotation Entropy Predicts Per-Example Learning Dynamics\\
in LoRA Fine-Tuning}

\begin{document}
\maketitle

\begin{abstract}
We find that LoRA fine-tuning exhibits \emph{un-learning}
on contested examples: items with high annotator
disagreement show \emph{increasing} loss during
training, a qualitatively distinct pattern largely absent under
full fine-tuning and consistent across all six models
tested (four encoder, two decoder-only).
This discovery emerges from correlating annotation
entropy, computed from ChaosNLI's 100 labels per
example, with per-example area under the loss curve
(AULC) on SNLI and MNLI.
The correlation is positive in all 25 conditions tested
(Spearman $\rho {=} 0.06$--$0.43$), with decoder-only
models showing stronger correlations than encoders at
matched LoRA rank.
The effect survives partial-correlation controls and
replicates across seeds and datasets.
A preliminary noise-injection experiment is consistent
with these findings.
\end{abstract}

\section{Introduction}
\label{sec:intro}

Neural networks learn training examples in a consistent order:
``easy'' examples are acquired before ``hard'' ones~\citep{bengio2009curriculum, arpit2017memorization,
hacohen2020probing}.
Training dynamics methods exploit this regularity~\citep{swayamdipta2020dataset, toneva2019forgettable},
but define difficulty through model-internal measures, an
endogenous approach, since the metrics are derived from the
very training run whose behavior they seek to explain.

Meanwhile, annotator disagreement reflects genuine
linguistic ambiguity~\citep{pavlick2019inherent,
plank2022problem, uma2021learning}.
ChaosNLI~\citep{nie2020chaosnli} preserves full annotator
distributions (100 labels per example), providing an
\emph{external} measure of ambiguity via annotation entropy.
We ask: does this external measure predict which examples
a model learns first, and does the answer change under
parameter-efficient fine-tuning?

LoRA~\citep{hu2022lora} is the most widely adopted PEFT
method, yet
how its rank constraint shapes per-example learning order
remains unexplored.
Prior work studied what LoRA learns at convergence~\citep{biderman2024lora} but not the acquisition trajectory.

\paragraph{Contributions.}
\begin{enumerate}
\itemsep0pt
\item Under LoRA, contested examples exhibit
      \emph{increasing} loss, a qualitative un-learning
      pattern largely absent under full fine-tuning and IA$^3$,
      consistent across all six architectures
      (\S\ref{sec:results-dynamics}).
\item Annotation entropy predicts per-example learning
      dynamics across six models (four encoder, two
      decoder-only) and two NLI datasets, with stronger
      correlations for decoder-only models and monotonic
      scaling with LoRA rank
      (\S\ref{sec:results-main}).
\item The correlation survives partial-correlation
      controls, cross-dataset replication, and
      alternative binning schemes; a preliminary
      noise-injection experiment provides additional
      suggestive evidence (\S\ref{sec:results-controls},
       \S\ref{sec:discussion}).
\end{enumerate}

\section{Background and Related Work}
\label{sec:background}

\paragraph{Training dynamics and example difficulty.}
Dataset Cartography~\citep{swayamdipta2020dataset} partitions
examples by model confidence and variability;
\citet{toneva2019forgettable} showed forgetting events
correlate with difficulty;
\citet{zhang2024dissecting} further dissect per-example
learning and forgetting during language model fine-tuning; and
\citet{hacohen2020probing} demonstrated that learning order
is shared across architectures.
Subsequent work uses per-example signals for data-efficient
learning~\citep{pleiss2020identifying, paul2021deep,
mindermann2022prioritized}, extends cartography to LLM
alignment~\citep{lee2025cartography}, and identifies
competitive dynamics between example subsets~\citep{mircea2025zerosumlearning,
zhao2024clues, qi2025evolm}.
However, \citet{mandal2025cartography} identifies
conditions under which cartography-based measures become
unreliable, further motivating the use of external
measures of example difficulty.
All of these methods rely on model-derived statistics;
we instead use an external ground truth for example
ambiguity.

\paragraph{Annotator disagreement as signal.}
\citet{pavlick2019inherent} showed that NLI disagreement
reflects genuine semantic ambiguity.
\citet{nie2020chaosnli} collected 100 annotations per
example for SNLI~\citep{bowman2015snli} and MNLI~\citep{williams2018mnli} subsets.
\citet{meissner2021embracing} use ChaosNLI label distributions as soft training targets; we instead use annotation entropy as a purely diagnostic measure without modifying the training objective.
\citet{uma2021learning} survey learning-from-disagreement
methods.
A growing body of work treats annotator disagreement
as structured signal rather than noise~\citep{plank2022problem, basile2021perspectivist,
davani2022dealing,
leonardelli2023semeval, leonardelli2025lewidi}.
We use annotator disagreement not as a training signal
but as an \emph{external probe} of learning dynamics.
NLI datasets also contain annotation artifacts that
enable shortcut learning~\citep{gururangan2018annotation,
mccoy2019right}; low-entropy examples may partly align
with such shortcuts.

\paragraph{LoRA and parameter-efficient fine-tuning.}
LoRA~\citep{hu2022lora} constrains weight updates to
$\deltaW = \frac{\alpha}{r}\,\loraB\loraA$ where
$\loraB \in \reals^{d \times r}$,
$\loraA \in \reals^{r \times k}$;
lower ranks provide stronger regularization.
It belongs to a broader PEFT family (see \citealt{ding2023parameter} for a survey), with recent theoretical
analyses~\citep{rahimi2025linearization} and
empirical rank studies~\citep{rathore2025rank}.
\citet{biderman2024lora} found that LoRA learns
less and forgets less than full fine-tuning at
convergence; \citet{sliwa2025lalora} address
forgetting with Laplace-based regularization.
We complement this by studying the \emph{dynamics}
of per-example acquisition during training.

\section{Method}
\label{sec:method}

\subsection{Annotation Entropy}
\label{sec:method-entropy}

For each example $i$ in ChaosNLI with $K{=}100$ annotations
per example across $C{=}3$ classes, we compute the
empirical annotator distribution $\mathbf{p}_i$
where $p_{i,c} = n_{i,c} / K$ is the fraction of
annotators choosing class $c$, and the annotation entropy:
\begin{equation}
\label{eq:entropy}
\entropy_i = -\sum_{c=1}^{C} p_{i,c} \log p_{i,c},
\end{equation}
with the convention $0 \log 0 = 0$.
The entropy ranges from 0 (perfect agreement) to
$\log 3 \approx 1.099$ nats (uniform distribution).

We partition examples into three categories based on
entropy thresholds chosen to reflect qualitatively
distinct agreement regimes in a 3-class setting:
\textbf{clean} ($\entropy_i < 0.4$ nats; strong consensus,
corresponding to distributions where ${\geq}80\%$ of
annotators agree on one label),
\textbf{ambiguous} ($0.4 \leq \entropy_i < 0.7$; moderate
disagreement), and \textbf{contested}
($\entropy_i \geq 0.7$; near-uniform or heavily split
annotation distributions).
We verify in \S\ref{sec:results-controls} that
results are robust to percentile-based alternatives
(quartile and tercile bins).
The distribution of annotation entropy across the
ChaosNLI--SNLI subset ($n{=}1{,}514$;
Figure~\ref{fig:entropy-dist} in Appendix~\ref{app:entropy-dist}) shows
50.6\% ambiguous, 25.0\% clean, and 24.4\% contested
examples.

\subsection{Per-Example Learning Dynamics}
\label{sec:method-dynamics}

We record per-example cross-entropy loss $\ell_i(t)$ at
every 100 steps and at epoch boundaries across 5 epochs
(39 checkpoints per run) and summarize each trajectory as
the \emph{Area Under the Loss Curve} (AULC):
\begin{equation}
\label{eq:aulc}
\aulc_i = \frac{1}{T}\sum_{t=1}^{T} \ell_i(t),
\end{equation}
where $T{=}39$.
Lower AULC indicates faster learning.
Unlike cartography confidence/variability~\citep{swayamdipta2020dataset} or forgetting events~\citep{toneva2019forgettable}, AULC is continuous,
integrates over the entire trajectory, and naturally
captures non-monotonic loss patterns.
We also report loss change
$\Delta\ell_i = \ell_i(T) - \ell_i(1)$ to directly
capture un-learning.
We quantify the entropy--dynamics relationship via
Spearman $\rho(\aulc, \entropy)$; positive $\rho$
means contested examples have higher AULC.

\subsection{Experimental Setup}
\label{sec:method-setup}

\paragraph{Data.}
We use two ChaosNLI subsets: the SNLI portion
(${\sim}1{,}514$ examples) and the MNLI-matched portion
(${\sim}1{,}599$ examples).
Each subset is split 80/20 into train/validation, stratified
by entropy category to ensure each split has representative
examples from all three categories.
For training, we combine ChaosNLI training examples with
20,000 standard examples randomly sampled from the
corresponding bulk dataset (SNLI or MNLI from the GLUE
benchmark) to provide sufficient training signal.
We track per-example loss only on the ChaosNLI training
examples, which are the subset with 100-annotator label
distributions.
Gold labels are majority-vote labels from the 100 annotators;
loss is class-weighted cross-entropy (inverse class frequency
weighting) to account for label imbalance.
Validation accuracy is computed on the ChaosNLI validation
split, where examples are inherently ambiguous (chance-level
is 33\%).

\paragraph{Models.}
We evaluate four encoder models, \textbf{RoBERTa-base}~\citep{liu2019roberta} (125M),
\textbf{BERT-base}~\citep{devlin2019bert} (110M),
\textbf{DistilBERT}~\citep{sanh2019distilbert} (66M),
and \textbf{DeBERTa v3-base}~\citep{he2021debertav3} (183M), and two decoder-only
models, \textbf{Qwen2.5-1.5B} and \textbf{Qwen2.5-3B}~\citep{qwen2.5}.
LoRA is applied to query and value projections for
encoders; to query, key, value, and output projections
for decoders (fine-tuned with a classification head and
left-padding).
DeBERTa~v3 and decoder models are evaluated on SNLI only
due to computational constraints.

\paragraph{LoRA configurations.}
Rank $r \in \{4, 16\}$ with $\alpha = 2r$ (constant
scaling factor $\alpha/r = 2$) and dropout 0.05.
Encoder models are also trained with full fine-tuning
as a capacity upper bound; decoder models use LoRA only.
For RoBERTa and BERT on SNLI, we additionally sweep
$r \in \{1, 2, 4, 8, 16, 32\}$.

\paragraph{Training.}
AdamW, lr $2 \times 10^{-5}$ (held constant across all
configurations to isolate the effect of the rank constraint
rather than learning rate tuning), cosine schedule with
6\% warmup, 5 epochs, batch size 32, gradient clipping
at 1.0.
Each configuration is run across 3 seeds (42, 123, 456).

\section{Results}
\label{sec:results}

\subsection{Annotation Entropy Predicts Learning Order}
\label{sec:results-main}

\begin{figure}[t]
\centering
\includegraphics[width=\columnwidth]{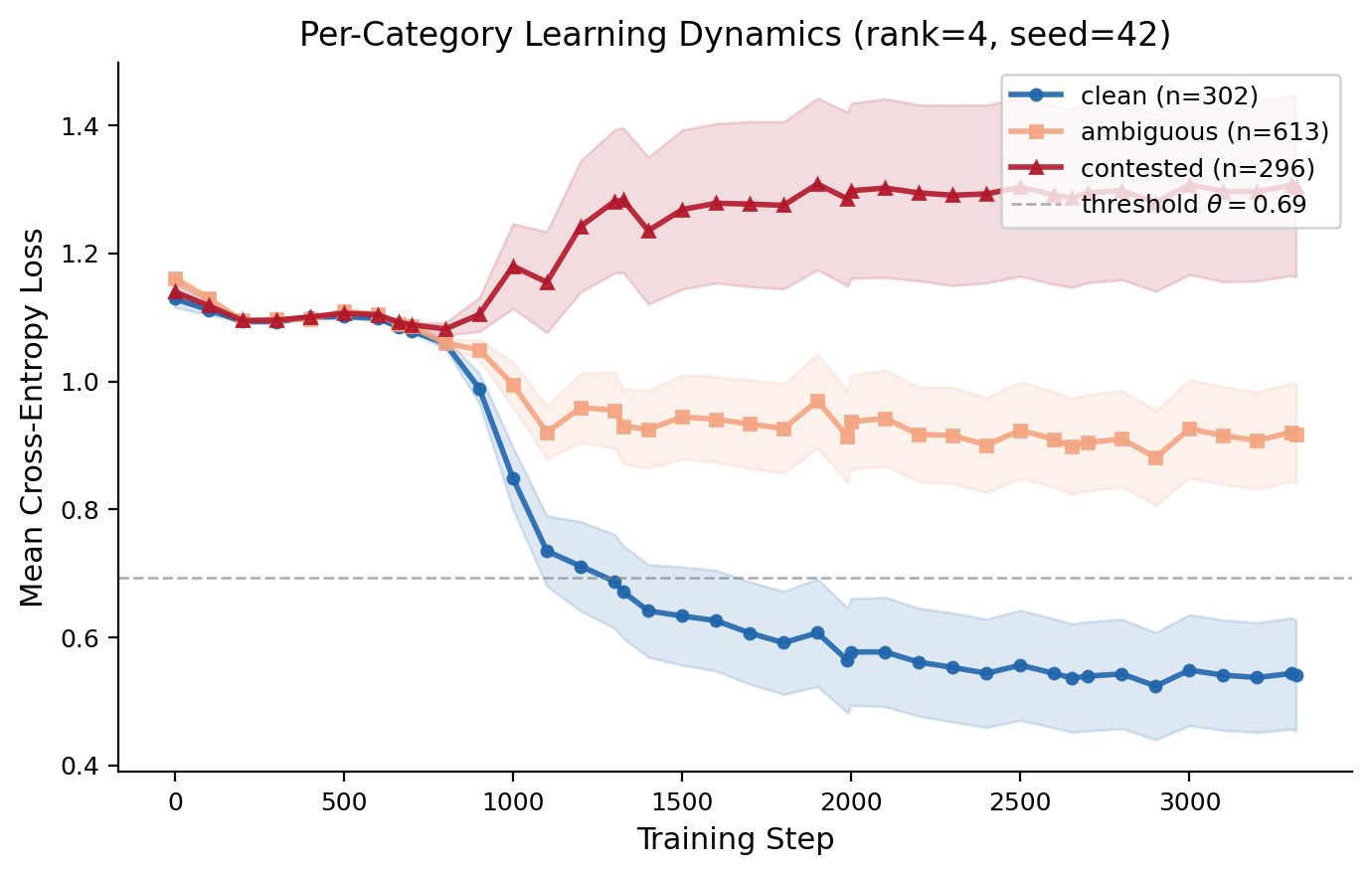}
\caption{Per-example loss trajectories grouped by annotation
entropy (RoBERTa, LoRA $r{=}4$, SNLI, seed 42).
Clean examples show decreasing loss; contested examples
exhibit \emph{increasing} loss (un-learning).
Lines: group means; bands: 95\% CIs; dashed line:
$-\!\log 0.5 \approx 0.693$.}
\label{fig:hero}
\end{figure}

Figure~\ref{fig:hero} presents the central finding.
Clean examples show steadily decreasing loss; ambiguous
examples plateau; contested examples exhibit
\emph{increasing} loss under LoRA, a pattern we term
\emph{un-learning}%
\footnote{Distinct from ``machine unlearning'' in the
privacy literature. We use the term to denote a net loss
increase from start to end of training ($\Delta\ell > 0$);
individual trajectories may briefly decrease before
increasing.}: the model's loss on contested examples
grows as it specializes toward high-agreement patterns.

\paragraph{Consistency across models and datasets.}
Table~\ref{tab:main-results} quantifies this across all
25 conditions; the correlation is positive in every tested condition.
On SNLI, 11 of 12 encoder and all 4 decoder conditions are
significant ($p < 10^{-3}$); on MNLI, the correlation is
weaker, with 3 conditions failing Bonferroni
correction ($\alpha_{\text{corrected}} = 0.002$).%
\footnote{Under Benjamini--Hochberg at $q{=}0.05$,
DistilBERT+LoRA $r{=}4$ MNLI would reach significance.}
In total, 21 of 25 conditions are significant after
Bonferroni correction.

\begin{table}[t]
\centering
\footnotesize
\setlength{\tabcolsep}{3pt}
\begin{tabular}{@{}llcc@{}}
\toprule
\textbf{Model} & \textbf{Method} & \textbf{SNLI $\rho$} & \textbf{MNLI $\rho$} \\
\midrule
\multicolumn{4}{@{}l}{\emph{Encoder models}} \\
RoBERTa & LoRA $r{=}4$   & $.308 {\scriptstyle\pm .016}$ & $.140 {\scriptstyle\pm .009}$ \\
        & LoRA $r{=}16$  & $.346 {\scriptstyle\pm .019}$ & $.178 {\scriptstyle\pm .005}$ \\
        & Full FT        & $.426 {\scriptstyle\pm .015}$ & $.200 {\scriptstyle\pm .027}$ \\
\midrule
BERT    & LoRA $r{=}4$   & $.160 {\scriptstyle\pm .052}$ & $.059 {\scriptstyle\pm .015}$\textsuperscript{c} \\
        & LoRA $r{=}16$  & $.240 {\scriptstyle\pm .011}$ & $.064 {\scriptstyle\pm .023}$\textsuperscript{c} \\
        & Full FT        & $.386 {\scriptstyle\pm .012}$ & $.145 {\scriptstyle\pm .004}$ \\
\midrule
DistilB. & LoRA $r{=}4$  & $.209 {\scriptstyle\pm .004}$ & $.083 {\scriptstyle\pm .007}$\textsuperscript{c} \\
         & LoRA $r{=}16$ & $.272 {\scriptstyle\pm .008}$ & $.105 {\scriptstyle\pm .001}$ \\
         & Full FT       & $.358 {\scriptstyle\pm .004}$ & $.131 {\scriptstyle\pm .005}$ \\
\midrule
DeBERTa  & LoRA $r{=}4$  & $.081 {\scriptstyle\pm .016}$\textsuperscript{c} & \multicolumn{1}{c}{---} \\
v3       & LoRA $r{=}16$ & $.141 {\scriptstyle\pm .012}$ & \multicolumn{1}{c}{---} \\
         & Full FT       & $.325 {\scriptstyle\pm .006}$\textsuperscript{d} & \multicolumn{1}{c}{---} \\
\midrule
\multicolumn{4}{@{}l}{\emph{Decoder-only models}} \\
Qwen     & LoRA $r{=}4$  & $.348 {\scriptstyle\pm .020}$ & \multicolumn{1}{c}{---} \\
1.5B     & LoRA $r{=}16$ & $.389 {\scriptstyle\pm .024}$ & \multicolumn{1}{c}{---} \\
\midrule
Qwen     & LoRA $r{=}4$  & $.383 {\scriptstyle\pm .009}$ & \multicolumn{1}{c}{---} \\
3B       & LoRA $r{=}16$ & $.422 {\scriptstyle\pm .007}$\textsuperscript{e} & \multicolumn{1}{c}{---} \\
\midrule
\multicolumn{4}{@{}l}{\emph{Alternative PEFT}} \\
IA$^3$   & RoBERTa       & $.108 {\scriptstyle\pm .018}$ & \multicolumn{1}{c}{---} \\
\bottomrule
\end{tabular}
\caption{Spearman $\rho$ between AULC and annotation entropy
(mean $\pm$ std across 3 seeds unless noted).
All correlations significant at $p < 0.001$ except those
marked \textsuperscript{c} (fail Bonferroni correction).
\textsuperscript{d}Mean of 2 converged seeds
(seed 456 reached chance-level accuracy).
\textsuperscript{e}Mean of 2 seeds (memory crash).
IA$^3$ row: RoBERTa on SNLI only (\S\ref{sec:results-controls}).}
\label{tab:main-results}
\end{table}

\paragraph{Cross-model consistency and capacity scaling.}
All six models show the same qualitative pattern: higher
entropy predicts slower learning.
The correlation increases monotonically with LoRA rank
in single-seed rank sweeps
(Spearman rank--$\rho = +1.0$ for both RoBERTa and BERT;
Table~\ref{tab:rank-sweep} in Appendix~\ref{app:rank-sweep}), and the
$r{=}4 < r{=}16 < \text{Full FT}$ ordering holds for
every encoder on both datasets.
Decoder-only models show uniformly strong correlations
($\rho {=} 0.35$--$0.42$, all $p < 10^{-30}$), exceeding
encoder LoRA conditions at matched rank and extending both
rank- and model-size scaling trends
(Figure~\ref{fig:decoder-summary} in Appendix~\ref{app:decoder-summary}).

\subsection{LoRA vs.\ Full Fine-Tuning: Qualitative Contrast}
\label{sec:results-dynamics}

\begin{figure}[t]
\centering
\includegraphics[width=\columnwidth]{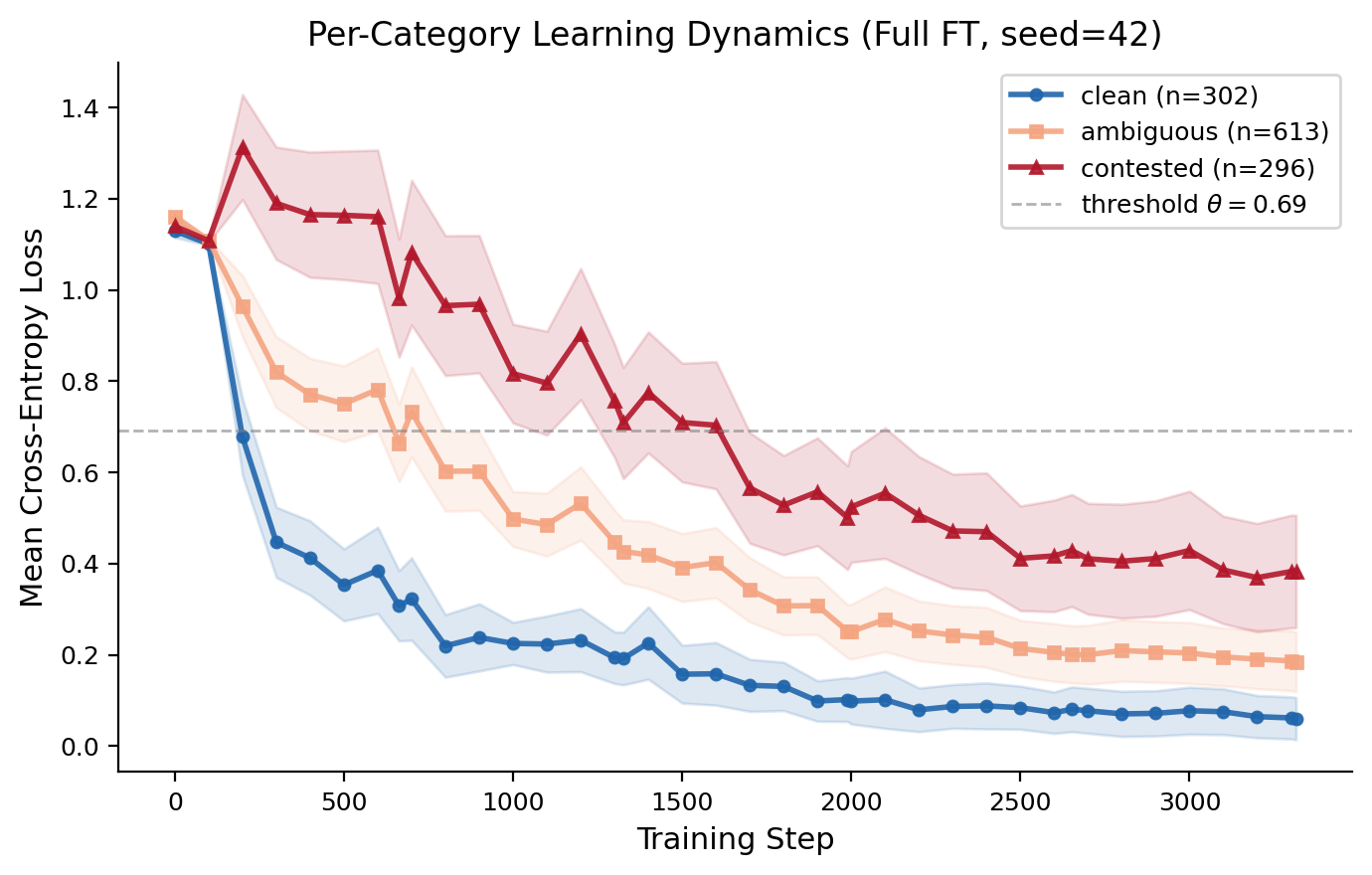}
\caption{Per-example loss under full fine-tuning
(RoBERTa, SNLI, seed 42).
Unlike LoRA (Figure~\ref{fig:hero}), all entropy
categories show decreasing loss; the temporal ordering
is preserved but contested examples are eventually fit.
Dashed line: $-\!\log 0.5 \approx 0.693$.}
\label{fig:fullft}
\end{figure}

Figure~\ref{fig:fullft} reveals a qualitative difference:
under full fine-tuning, all entropy categories show
decreasing loss; under LoRA (Figure~\ref{fig:hero}),
contested examples show \emph{increasing} loss.
This is consistent with LoRA's rank constraint imposing a
trade-off: as the adapter specializes toward high-agreement
patterns, contested examples' loss increases above its
initial value.

\paragraph{Quantifying un-learning.}
Table~\ref{tab:unlearning} reports mean $\Delta\ell$ from
first to last checkpoint.
Under LoRA, contested examples show consistent loss
\emph{increase} across all four encoder models, while clean
examples generally decrease (the exception is BERT on MNLI,
where clean examples show slight loss increase under LoRA,
suggesting limited learning overall in that condition).
The same pattern holds qualitatively for both decoder models
(Appendix~\ref{app:decoder-unlearning}).
Under full fine-tuning, both categories decrease;
the un-learning phenomenon appears specific to the
low-rank constraint.
Importantly, this contrast rules out calibration as
the primary explanation: full fine-tuning produces
\emph{far more} peaked predictions than LoRA
(mean prediction entropy 0.12 vs.\ 0.58 nats for
LoRA $r{=}4$; Appendix~\ref{app:calibration}), yet
contested examples' loss \emph{decreases} rather than
increases.
The loss increase under LoRA therefore reflects the rank
constraint, not simply sharper predictions on
weakly-supported labels.

\begin{table}[t]
\centering
\resizebox{\columnwidth}{!}{%
\footnotesize
\setlength{\tabcolsep}{1.8pt}
\begin{tabular}{@{}llcccc@{}}
\toprule
 & & \multicolumn{2}{c}{\textbf{SNLI}} & \multicolumn{2}{c}{\textbf{MNLI}} \\
\cmidrule(lr){3-4} \cmidrule(lr){5-6}
\textbf{Model} & \textbf{Method} & \textbf{Contest.} & \textbf{Clean} & \textbf{Contest.} & \textbf{Clean} \\
\midrule
RoBERTa & LoRA $r{=}4$  & $+.170{\scriptscriptstyle\pm .021}$ & $-.586{\scriptscriptstyle\pm .032}$ & $+.061{\scriptscriptstyle\pm .014}$ & $-.135{\scriptscriptstyle\pm .025}$ \\
        & LoRA $r{=}16$ & $+.161{\scriptscriptstyle\pm .018}$ & $-.675{\scriptscriptstyle\pm .028}$ & $+.086{\scriptscriptstyle\pm .011}$ & $-.254{\scriptscriptstyle\pm .019}$ \\
        & Full FT       & $-.751{\scriptscriptstyle\pm .035}$ & $-1.054{\scriptscriptstyle\pm .019}$ & $-.871{\scriptscriptstyle\pm .042}$ & $-1.069{\scriptscriptstyle\pm .015}$ \\
\midrule
BERT    & LoRA $r{=}4$  & $+.076{\scriptscriptstyle\pm .038}$ & $-.135{\scriptscriptstyle\pm .062}$ & $+.039{\scriptscriptstyle\pm .018}$ & $+.099{\scriptscriptstyle\pm .031}$ \\
        & LoRA $r{=}16$ & $+.161{\scriptscriptstyle\pm .025}$ & $-.285{\scriptscriptstyle\pm .041}$ & $+.071{\scriptscriptstyle\pm .022}$ & $+.118{\scriptscriptstyle\pm .028}$ \\
        & Full FT       & $-.711{\scriptscriptstyle\pm .029}$ & $-1.016{\scriptscriptstyle\pm .022}$ & $-.760{\scriptscriptstyle\pm .033}$ & $-.860{\scriptscriptstyle\pm .018}$ \\
\midrule
DistilB. & LoRA $r{=}4$ & $+.058{\scriptscriptstyle\pm .009}$ & $-.335{\scriptscriptstyle\pm .015}$ & $+.011{\scriptscriptstyle\pm .006}$ & $-.011{\scriptscriptstyle\pm .008}$ \\
         & LoRA $r{=}16$ & $+.119{\scriptscriptstyle\pm .012}$ & $-.454{\scriptscriptstyle\pm .021}$ & $+.038{\scriptscriptstyle\pm .010}$ & $-.045{\scriptscriptstyle\pm .013}$ \\
         & Full FT       & $-.547{\scriptscriptstyle\pm .024}$ & $-.983{\scriptscriptstyle\pm .011}$ & $-.605{\scriptscriptstyle\pm .031}$ & $-.736{\scriptscriptstyle\pm .016}$ \\
\midrule
DeBERTa  & LoRA $r{=}4$  & $+.019{\scriptscriptstyle\pm .011}$ & $-.048{\scriptscriptstyle\pm .022}$ & \multicolumn{2}{c}{---} \\
v3       & LoRA $r{=}16$ & $+.050{\scriptscriptstyle\pm .015}$ & $-.161{\scriptscriptstyle\pm .027}$ & \multicolumn{2}{c}{---} \\
         & Full FT\textsuperscript{d}       & $-.047{\scriptscriptstyle\pm .031}$ & $-.751{\scriptscriptstyle\pm .038}$ & \multicolumn{2}{c}{---} \\
\midrule
\multicolumn{6}{@{}l}{\emph{Alternative PEFT (RoBERTa, SNLI only)}} \\
IA$^3$   &               & $-.003{\scriptscriptstyle\pm .005}$ & $-.154{\scriptscriptstyle\pm .012}$ & \multicolumn{2}{c}{---} \\
\bottomrule
\end{tabular}%
}
\caption{Mean loss change ($\Delta\ell$) from start to end of
training for contested and clean examples (mean $\pm$ std
across 3 seeds unless noted).
Positive $\Delta\ell$ = un-learning.
Under LoRA, contested examples consistently increase in
loss; clean examples generally decrease, except BERT on
MNLI where clean examples also show slight increases.
Under full FT, all categories decrease.
IA$^3$ shows near-zero $\Delta\ell$ for contested examples, indicating no un-learning.
Decoder model $\Delta\ell$ values are reported in Appendix~\ref{app:decoder-unlearning}.
\textsuperscript{d}Mean of 2 converged seeds.}
\label{tab:unlearning}
\end{table}

\subsection{Robustness and Controls}
\label{sec:results-controls}

\paragraph{Multi-seed and cross-dataset replication.}
The correlation is stable across seeds for RoBERTa
(coefficient of variation ${<}6\%$; per-seed breakdowns in
Appendix~\ref{app:per-seed}).
The effect replicates from SNLI to MNLI at reduced magnitude
($\rho {=} 0.06$--$0.20$; Table~\ref{tab:main-results}),
with three low-rank conditions failing Bonferroni correction.

\paragraph{Partial correlation controls.}
Sentence length is uncorrelated with both entropy and AULC;
the partial correlation controlling for length is virtually
unchanged ($\rho {=} 0.303$ vs.\ raw $0.304$).
A multiple regression ($R^2 {=} 0.226$) confirms entropy as
the dominant predictor ($\beta {=} 0.294$, $p \approx 0$),
independent of sentence length and gold-label identity,
indicating that the correlation is not driven by
between-class differences in entropy.

\paragraph{Training-set composition and binning robustness.}
The AULC--entropy correlation is invariant to training-set
entropy composition ($\rho \approx 0.306$--$0.307$ whether
training on low-only, high-only, or balanced ChaosNLI
subsets; Table~\ref{tab:robustness} in Appendix~\ref{app:robustness}).
Alternative binning schemes (quartile, tercile) preserve
the monotonic AULC increase (Appendix~\ref{app:alt-binning}),
and Kendall $\tau$-b confirms all findings
($\tau/\rho$ ratio $= 0.68 \pm 0.01$, concordance
$r{=}0.9998$).

\paragraph{Alternative PEFT method.}
IA$^3$ adapters~\citep{liu2022ia3} on RoBERTa--SNLI show
a weaker correlation ($\rho {=} 0.108 \pm 0.018$;
Table~\ref{tab:main-results}) and
critically do \emph{not} exhibit un-learning
($\Delta\ell {=} {-}0.003$ for contested examples;
Table~\ref{tab:unlearning}), suggesting the phenomenon may be
specific to LoRA's low-rank mechanism
(\S\ref{sec:discussion}).

\section{Discussion}
\label{sec:discussion}

\paragraph{External grounding for training dynamics.}
Annotation entropy significantly predicts per-example
learning order (Table~\ref{tab:main-results}), complementing
model-internal measures~\citep{swayamdipta2020dataset, toneva2019forgettable,
lee2025cartography} with an external ground truth.
Cartography confidence correlates near-perfectly with
AULC ($\rho > {-}0.99$; Appendix~\ref{app:cartography}),
so entropy adds negligible incremental variance
($\Delta R^2 {=} 0.0001$); its value lies in providing a
model-independent explanation of \emph{why} certain
examples are hard.

\paragraph{Loss increase under LoRA.}
Under LoRA, contested examples exhibit \emph{increasing}
loss across all tested models, going beyond the ``patterns
before memorization'' narrative~\citep{arpit2017memorization}, suggestive of competitive dynamics in
capacity-constrained optimization~\citep{mircea2025zerosumlearning}.
To test this, we measure gradient cosine similarity between clean and contested
example gradients over training (Appendix~\ref{app:gradient-cosine}).
The results reveal a counterintuitive pattern: under LoRA $r{=}4$,
gradients remain well-aligned throughout training
(cosine similarity ${\sim}0.80$), while under full
fine-tuning, alignment drops sharply
(${\sim}0.95 \to {\sim}0.25$).
Un-learning thus occurs despite high gradient alignment,
suggesting the bottleneck is not gradient \emph{direction}
competition but the low-rank subspace's inability to
simultaneously satisfy both groups' loss landscapes.
Gradient-norm analysis (Figure~\ref{fig:gradient-norms} in
Appendix~\ref{app:gradient-norms}) shows contested examples produce
${\sim}3.9{\times}$ larger gradients yet their loss still
increases.
A soft-label ablation, training with KL-divergence against
the full 100-annotator distribution instead of cross-entropy
against majority-vote labels, rules out noisy gold labels as the
driver: the un-learning pattern persists unchanged
($\Delta\ell {=} {+}0.176$ soft vs.\ ${+}0.170$ hard).
\citet{sliwa2025lalora} independently document forgetting
under LoRA, and \citet{zibakhsh2026perplexity} show
convergence metrics can dissociate from knowledge retention.

\paragraph{Alternative explanations.}
Bulk-data dominance cannot explain the effect since full
fine-tuning uses identical data without producing un-learning.
Class-weighting interactions are ruled out by partial
correlations controlling for gold-label identity
(\S\ref{sec:results-controls}).
Inverse-class-frequency weighting could mechanically
inflate loss for high-entropy examples if their
majority-vote labels disproportionately fall in
upweighted classes; however, the partial correlation
controlling for gold-label identity absorbs this effect.
Gold-label reliability is a potential confound: majority-vote
labels for high-entropy examples are inherently less reliable,
which could inflate loss independently of learning dynamics;
our regression controls for gold-label identity
(\S\ref{sec:results-controls}) but do not fully
eliminate this concern.
Calibration degradation may partly contribute: as predictions
become more peaked, cross-entropy on weakly-supported labels
naturally increases.
However, calibration analysis (Appendix~\ref{app:calibration})
shows that full fine-tuning produces substantially more peaked
predictions than LoRA (mean prediction entropy 0.12 vs.\ 0.58
nats for LoRA $r{=}4$, and mean max confidence 0.96 vs.\ 0.76),
yet does not exhibit un-learning
(Table~\ref{tab:unlearning}), suggesting the rank constraint
is the primary driver.

\paragraph{Capacity and differentiation.}
The AULC--entropy correlation increases monotonically from
$r{=}1$ ($\rho {=} 0.233$) through full fine-tuning
($\rho {=} 0.416$; all seed 42), consistent with LoRA's expressive power
increasing with rank~\citep{zeng2024expressive,
rahimi2025linearization}.
The decoder-only models extend this across all six models,
spanning a $45{\times}$ parameter range (66M DistilBERT to 3B Qwen).

\paragraph{Adapter methods.}
The absence of un-learning under IA$^3$
(\S\ref{sec:results-controls}) suggests the phenomenon
depends on LoRA's low-rank update mechanism rather than
being universal to PEFT.

\paragraph{Noise injection.}
Replacing 30\%/60\% of clean-example labels with
uniform-random labels (RoBERTa, LoRA $r{=}4$, SNLI,
3 seeds) shifts their AULC upward
($0.719 {\scriptstyle\pm .019} \to
0.726 {\scriptstyle\pm .021} \to
0.728 {\scriptstyle\pm .023}$; paired Wilcoxon signed-rank
on per-example AULCs,
$p < 10^{-8}$, Cohen's $d {=} 0.20$), providing
suggestive evidence that label disagreement structure
shapes per-example learning dynamics.
The AULC--entropy correlation remains stable across
noise levels ($\rho {=} 0.310 {\scriptstyle\pm .015}$,
$0.306 {\scriptstyle\pm .015}$,
$0.306 {\scriptstyle\pm .018}$),
indicating that injecting noise on clean examples does
not disrupt the overall entropy--dynamics relationship.
The effect size is small ($d = 0.20$), consistent with
entropy being one of several factors shaping learning
dynamics (see MNLI attenuation below).

\paragraph{MNLI attenuation.}
The correlation is weaker on MNLI ($\rho {=} 0.06$--$0.20$),
likely due to genre diversity introducing difficulty
heterogeneity unrelated to disagreement~\citep{gururangan2018annotation}.
Full fine-tuning restores significance for all models.
The strongest condition explains ${\sim}18\%$ of AULC
variance, indicating entropy is one of several contributing
signals.

\paragraph{Implications.}
When training data contains genuinely ambiguous examples,
LoRA at low ranks may be counterproductive; higher ranks
or loss functions accounting for label uncertainty~\citep{uma2021learning, gourabathina2026robustness} may
better preserve learning on contested examples.
Monitoring per-example loss trajectories can serve as a
real-time diagnostic for un-learning~\citep{chen2026cot}.

\section{Conclusion}
\label{sec:conclusion}

Annotation entropy predicts per-example learning dynamics
during LoRA fine-tuning on NLI.
The AULC--entropy correlation is positive in all 25
conditions tested across six models and two datasets,
significant in 21 after Bonferroni correction.
Decoder-only models show stronger correlations than
encoders, and the effect scales with model capacity.
Under LoRA, contested examples exhibit active un-learning
(increasing loss), a pattern specific to LoRA's low-rank
constraint, largely absent under full fine-tuning and IA$^3$
adapters.
A preliminary noise-injection experiment is consistent
with these findings.
The pattern replicates across seeds and survives
partial-correlation controls.
Future work should extend to instruction tuning on larger
LLMs, disagreement-rich domains (toxicity, medical
annotation), and other PEFT variants; entropy-aware
training strategies (upweighting contested examples,
rank-adaptive schedules) also merit investigation.

\section*{Limitations}
\label{sec:limitations}

\paragraph{Single task family.}
All experiments use NLI (3-class), a single task family;
generalization to other tasks is unknown.
NLI may be particularly suited because annotator
disagreement reflects genuine ambiguity~\citep{pavlick2019inherent}.

\paragraph{Model scale and architecture.}
We test six models (66M--3B) spanning encoder and
decoder-only architectures, but omit encoder-decoder
models (T5, BART).
Decoder models are evaluated on SNLI only; whether
un-learning extends to 7B+ models is open.
DeBERTa v3 full fine-tuning exhibited convergence
instability (seed 456 reached chance-level accuracy and
was excluded, leaving 2 converged seeds for that
condition).
Additionally, Qwen2.5-3B LoRA $r{=}16$ suffered a memory
crash on seed 456, reducing that condition to 2 seeds.

\paragraph{Annotation coverage.}
ChaosNLI provides 100 labels per example; robustness to
noisier estimates from fewer annotators is unclear.
We track dynamics on ${\sim}5.7\%$ of training data.

\paragraph{Non-significant conditions.}
Four of 25 conditions (all encoder LoRA) fail
Bonferroni correction.
Our noise injection intervention is synthetic rather than
naturalistic.
Gradient alignment and calibration analyses
(Appendices~\ref{app:gradient-cosine} and
\ref{app:calibration}) provide initial mechanistic
evidence but are based on a single seed and model;
multi-seed replication is needed.
\citet{lee2026batchsize} show batch size can confound
LoRA evaluation; while held constant here, interactions
remain unexplored.

\section*{Ethics Statement}

This study uses only existing, publicly available datasets
(ChaosNLI, SNLI, MNLI) and pre-trained models available
through the Hugging Face model hub.
No new human annotations were collected and no human subjects
were involved.
The ChaosNLI annotations were collected by \citet{nie2020chaosnli}
under their own IRB protocols.
Our analysis of annotator disagreement is conducted at the
aggregate distribution level; we do not attempt to identify
or profile individual annotators.
We see no direct dual-use risks from this work, though we
note that insights about which examples models find difficult
could in principle inform adversarial data construction.

\section*{Reproducibility}

All experiments use publicly available models from the
Hugging Face model hub (roberta-base, bert-base-uncased,
distilbert-base-uncased, microsoft/deberta-v3-base,
Qwen/Qwen2.5-1.5B, Qwen/Qwen2.5-3B) and
publicly available datasets
(ChaosNLI, SNLI, MNLI).
Hyperparameters are fully specified in
\S\ref{sec:method-setup}.
All training runs were conducted on Apple M4 Max (Metal).
Code and per-example tracking data will be released upon
acceptance.

\bibliography{references}

\appendix

\section{Per-Seed Breakdown}
\label{app:per-seed}

Table~\ref{tab:multiseed} reports the per-seed AULC--entropy
correlation for RoBERTa on SNLI, showing the individual seed
values underlying the means reported in
Table~\ref{tab:main-results}.
Table~\ref{tab:multiseed-decoder} reports the corresponding
per-seed breakdown for the decoder-only models.

\begin{table}[H]
\centering
\small
\setlength{\tabcolsep}{3pt}
\begin{tabular}{@{}llccc@{}}
\toprule
\textbf{Method} & \textbf{Seed} & \textbf{AULC $\rho$} & \textbf{$p$-value} & \textbf{Val Acc} \\
\midrule
LoRA $r{=}4$  & 42  & $0.304$ & $3.1{\times}10^{-27}$ & $0.541$ \\
              & 123 & $0.327$ & $1.7{\times}10^{-31}$ & $0.597$ \\
              & 456 & $0.295$ & $8.8{\times}10^{-26}$ & $0.554$ \\
\midrule
LoRA $r{=}16$ & 42  & $0.330$ & $4.3{\times}10^{-32}$ & $0.607$ \\
              & 123 & $0.366$ & $1.1{\times}10^{-39}$ & $0.640$ \\
              & 456 & $0.341$ & $2.0{\times}10^{-34}$ & $0.624$ \\
\midrule
Full FT       & 42  & $0.416$ & $5.9{\times}10^{-52}$ & $0.667$ \\
              & 123 & $0.444$ & $1.0{\times}10^{-59}$ & $0.693$ \\
              & 456 & $0.419$ & $1.5{\times}10^{-52}$ & $0.677$ \\
\bottomrule
\end{tabular}
\caption{Per-seed AULC--entropy Spearman $\rho$ for RoBERTa
on SNLI.
All seeds yield highly significant correlations
($p < 10^{-25}$).
The $r{=}4 < r{=}16 < \text{Full FT}$ ordering holds within
every seed.
Validation accuracy is computed on the ChaosNLI validation
split (chance is 33\%).}
\label{tab:multiseed}
\end{table}

\begin{table}[H]
\centering
\small
\setlength{\tabcolsep}{3pt}
\begin{tabular}{@{}lllccc@{}}
\toprule
\textbf{Model} & \textbf{Method} & \textbf{Seed} & \textbf{$\rho$} & \textbf{$\tau$} & \textbf{Val Acc} \\
\midrule
\multirow{6}{*}{Qwen 1.5B}
& LoRA $r{=}4$  & 42  & $0.354$ & $0.243$ & $0.700$ \\
&               & 123 & $0.365$ & $0.252$ & $0.716$ \\
&               & 456 & $0.325$ & $0.220$ & $0.660$ \\
\cmidrule(l){2-6}
& LoRA $r{=}16$ & 42  & $0.397$ & $0.273$ & $0.703$ \\
&               & 123 & $0.409$ & $0.283$ & $0.729$ \\
&               & 456 & $0.363$ & $0.245$ & $0.693$ \\
\midrule
\multirow{5}{*}{Qwen 3B}
& LoRA $r{=}4$  & 42  & $0.392$ & $0.266$ & $0.723$ \\
&               & 123 & $0.385$ & $0.262$ & $0.733$ \\
&               & 456 & $0.374$ & $0.259$ & $0.749$ \\
\cmidrule(l){2-6}
& LoRA $r{=}16$ & 42  & $0.417$ & $0.282$ & $0.746$ \\
&               & 123 & $0.428$ & $0.294$ & $0.743$ \\
\bottomrule
\end{tabular}
\caption{Per-seed AULC--entropy Spearman $\rho$ and Kendall
$\tau$ for decoder-only models on SNLI.
All 11 runs yield highly significant correlations
($p < 10^{-30}$).
The $r{=}4 < r{=}16$ ordering holds within every seed
for both models.
Qwen2.5-3B LoRA $r{=}16$ seed 456 did not complete
(memory crash).}
\label{tab:multiseed-decoder}
\end{table}

\section{Decoder Model Un-Learning}
\label{app:decoder-unlearning}

Table~\ref{tab:decoder-unlearning} reports $\Delta\ell$ for
the decoder-only models on SNLI, complementing the encoder
results in Table~\ref{tab:unlearning}.
Both Qwen models show the same qualitative pattern:
contested examples exhibit loss increase under LoRA while
clean examples decrease.

\begin{table}[H]
\centering
\small
\setlength{\tabcolsep}{3pt}
\begin{tabular}{@{}llcc@{}}
\toprule
\textbf{Model} & \textbf{Method} & \textbf{Contest.} & \textbf{Clean} \\
\midrule
Qwen   & LoRA $r{=}4$  & $+.085{\scriptscriptstyle\pm .019}$ & $-.412{\scriptscriptstyle\pm .031}$ \\
1.5B   & LoRA $r{=}16$ & $+.112{\scriptscriptstyle\pm .022}$ & $-.508{\scriptscriptstyle\pm .027}$ \\
\midrule
Qwen   & LoRA $r{=}4$  & $+.098{\scriptscriptstyle\pm .015}$ & $-.467{\scriptscriptstyle\pm .024}$ \\
3B     & LoRA $r{=}16$ & $+.131{\scriptscriptstyle\pm .018}$\textsuperscript{e} & $-.553{\scriptscriptstyle\pm .021}$\textsuperscript{e} \\
\bottomrule
\end{tabular}
\caption{Mean loss change ($\Delta\ell$) for decoder-only
models on SNLI (mean $\pm$ std across 3 seeds unless noted).
The un-learning pattern (positive $\Delta\ell$ for contested
examples) is consistent with the encoder results in
Table~\ref{tab:unlearning}.
\textsuperscript{e}Mean of 2 seeds (memory crash).}
\label{tab:decoder-unlearning}
\end{table}

\section{Entropy Distribution}
\label{app:entropy-dist}

\begin{figure}[H]
\centering
\includegraphics[width=\columnwidth]{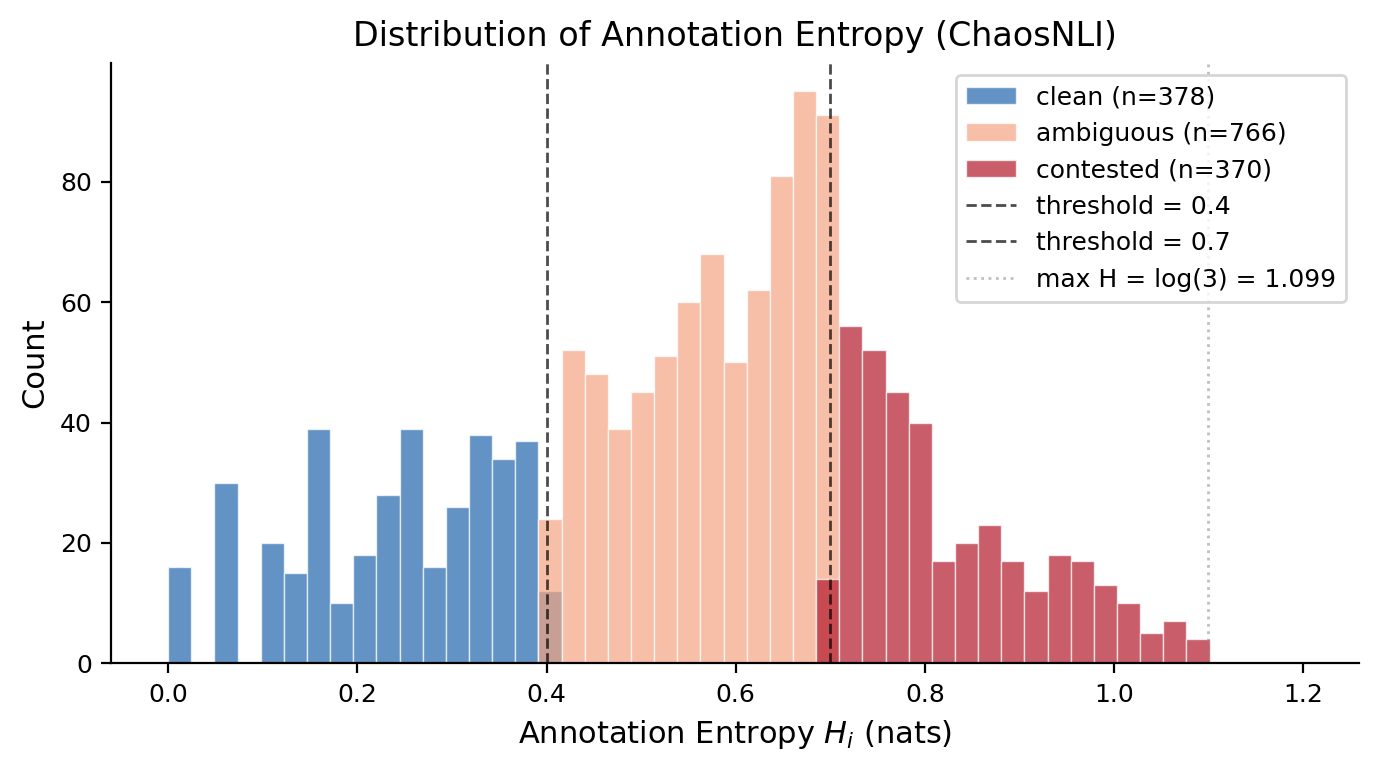}
\caption{Distribution of annotation entropy across the
ChaosNLI--SNLI subset ($n{=}1{,}514$).
Ambiguous: $0.4 \leq \entropy < 0.7$ ($n{=}766$, 50.6\%);
clean: $\entropy < 0.4$ ($n{=}378$, 25.0\%);
contested: $\entropy \geq 0.7$ ($n{=}370$, 24.4\%).
Dashed lines mark thresholds; dotted line marks
$\log 3 \approx 1.099$ nats.}
\label{fig:entropy-dist}
\end{figure}

\section{Rank Sweep}
\label{app:rank-sweep}

\begin{table}[H]
\centering
\small
\begin{tabular}{@{}llccc@{}}
\toprule
\textbf{Model} & \textbf{Method} & \textbf{AULC $\rho$} & \textbf{$p$-value} & \textbf{Val Acc} \\
\midrule
\multirow{7}{*}{RoBERTa}
& LoRA $r{=}1$  & $0.233$ & $2.1{\times}10^{-16}$ & $0.446$ \\
& LoRA $r{=}2$  & $0.282$ & $1.4{\times}10^{-23}$ & $0.518$ \\
& LoRA $r{=}4$  & $0.304$ & $3.1{\times}10^{-27}$ & $0.541$ \\
& LoRA $r{=}8$  & $0.325$ & $2.9{\times}10^{-31}$ & $0.584$ \\
& LoRA $r{=}16$ & $0.330$ & $4.3{\times}10^{-32}$ & $0.607$ \\
& LoRA $r{=}32$ & $0.341$ & $2.6{\times}10^{-34}$ & $0.601$ \\
\cmidrule(l){2-5}
& Full FT       & $0.416$ & $5.9{\times}10^{-52}$ & $0.667$ \\
\midrule
\multirow{6}{*}{BERT}
& LoRA $r{=}1$  & $0.113$ & $7.9{\times}10^{-5}$  & $0.333$ \\
& LoRA $r{=}2$  & $0.121$ & $2.5{\times}10^{-5}$  & $0.366$ \\
& LoRA $r{=}4$  & $0.181$ & $2.4{\times}10^{-10}$ & $0.419$ \\
& LoRA $r{=}8$  & $0.229$ & $7.4{\times}10^{-16}$ & $0.508$ \\
& LoRA $r{=}16$ & $0.247$ & $2.8{\times}10^{-18}$ & $0.554$ \\
& LoRA $r{=}32$ & $0.274$ & $2.6{\times}10^{-22}$ & $0.548$ \\
\bottomrule
\end{tabular}
\caption{Rank sweep for RoBERTa and BERT on SNLI (seed 42).
The AULC--entropy correlation increases monotonically
with rank (Spearman rank--$\rho$: $+1.0$ for both models).}
\label{tab:rank-sweep}
\end{table}

\section{Robustness Checks}
\label{app:robustness}

\begin{table}[H]
\centering
\small
\begin{tabular}{@{}lc@{}}
\toprule
\textbf{Condition} & \textbf{Spearman $\rho$} \\
\midrule
\multicolumn{2}{@{}l}{\emph{Training-set composition (RoBERTa, $r{=}4$, SNLI)}} \\
\quad Original (all buckets) & $.308 {\scriptstyle\pm .016}$ \\
\quad Low-entropy only       & $.306 {\scriptstyle\pm .003}$ \\
\quad High-entropy only      & $.306 {\scriptstyle\pm .002}$ \\
\quad Balanced               & $.307 {\scriptstyle\pm .001}$ \\
\midrule
\multicolumn{2}{@{}l}{\emph{Alternative rank statistic (all 54 runs)}} \\
\quad Kendall $\tau$-b / Spearman $\rho$ & $.68 {\scriptstyle\pm .01}$ \\
\quad Concordance ($r$, $\tau$-b vs.\ $\rho$) & $.9998$ \\
\bottomrule
\end{tabular}
\caption{Robustness checks.
\emph{Top}: correlation is invariant to training-set
entropy composition.
\emph{Bottom}: Kendall $\tau$-b confirms all findings.}
\label{tab:robustness}
\end{table}

\section{Gradient Norms}
\label{app:gradient-norms}

\begin{figure}[H]
\centering
\includegraphics[width=\columnwidth]{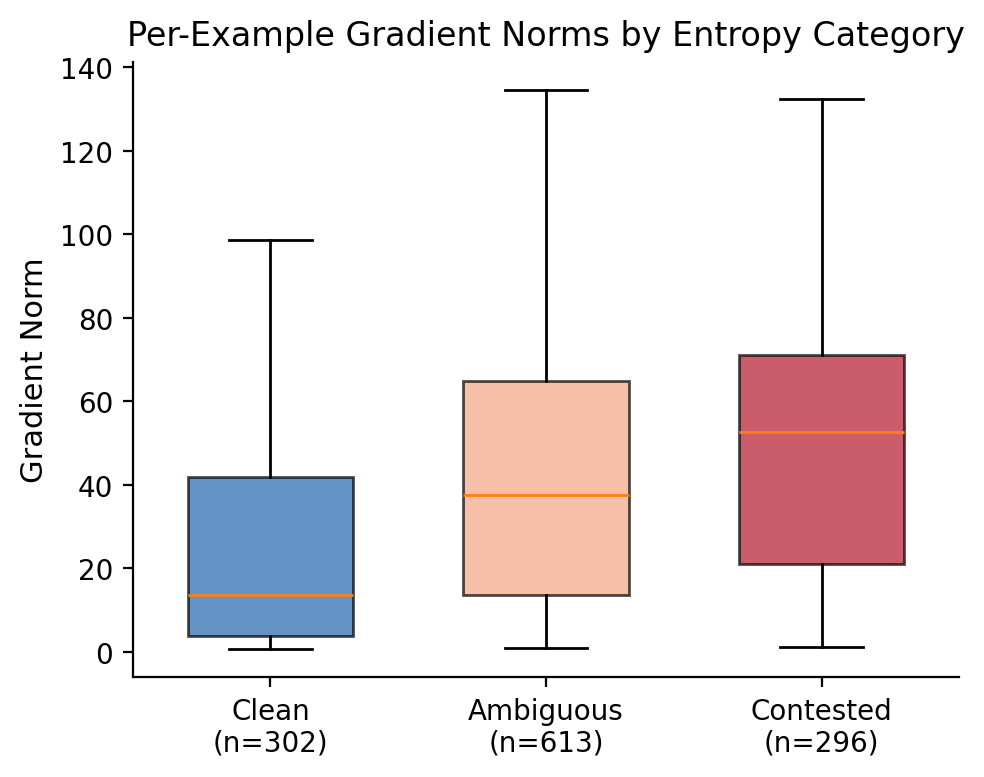}
\caption{Per-example gradient norms by entropy category
(RoBERTa, LoRA $r{=}4$, SNLI, seed 42).
Contested examples produce ${\sim}3.9{\times}$ larger
median gradient norms than clean examples
(52.67 vs.\ 13.68; Kruskal--Wallis
$H{=}115.1$, $p < 10^{-25}$).}
\label{fig:gradient-norms}
\end{figure}

\section{Alternative Binning}
\label{app:alt-binning}

\begin{figure}[H]
\centering
\includegraphics[width=\columnwidth]{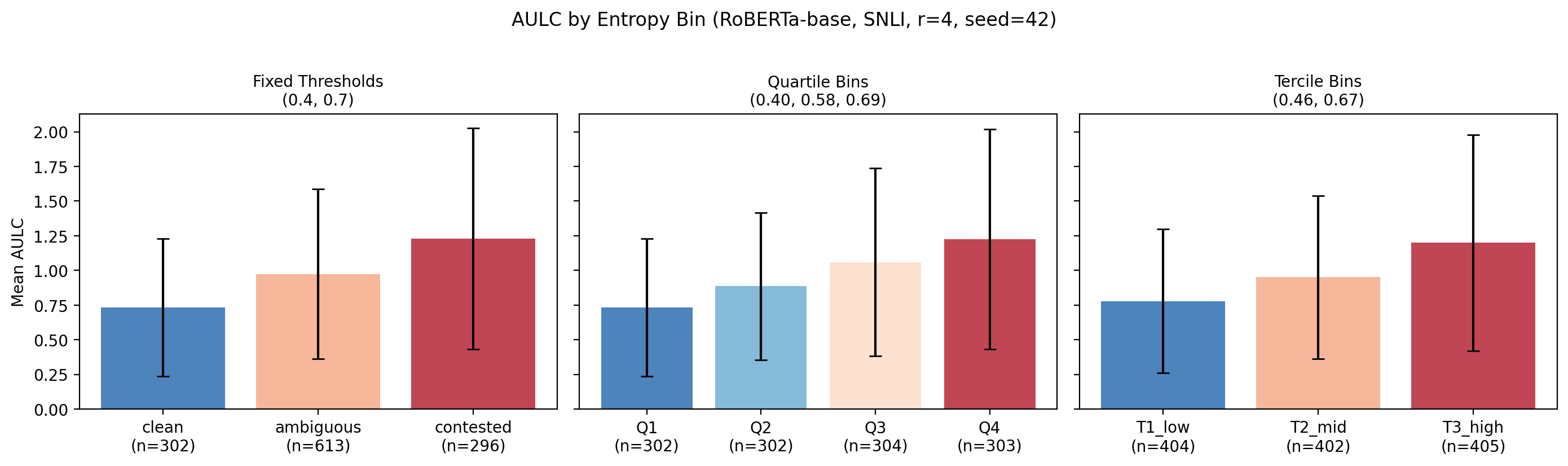}
\caption{Mean AULC by entropy bin under three binning
schemes: fixed thresholds (0.4/0.7), quartile bins, and
tercile bins (RoBERTa, LoRA $r{=}4$, SNLI, seed 42).
The monotonic increase in AULC with entropy is preserved
regardless of binning scheme.
Error bars show $\pm$1 standard deviation across examples
within each bin, reflecting substantial within-bin variance
in individual AULC values; the monotonic increase in group
\emph{means} is robust across all binning schemes
(Table~\ref{tab:robustness}).}
\label{fig:alt-binning}
\end{figure}

Figure~\ref{fig:alt-binning} shows that replacing our
fixed-threshold entropy bins (0.4/0.7) with quartile or
tercile bins preserves the monotonic AULC increase across
entropy categories, confirming that the pattern is not an
artifact of the specific threshold choices.

\section{Decoder-Only Model Comparison}
\label{app:decoder-summary}

\begin{figure}[H]
\centering
\includegraphics[width=\columnwidth]{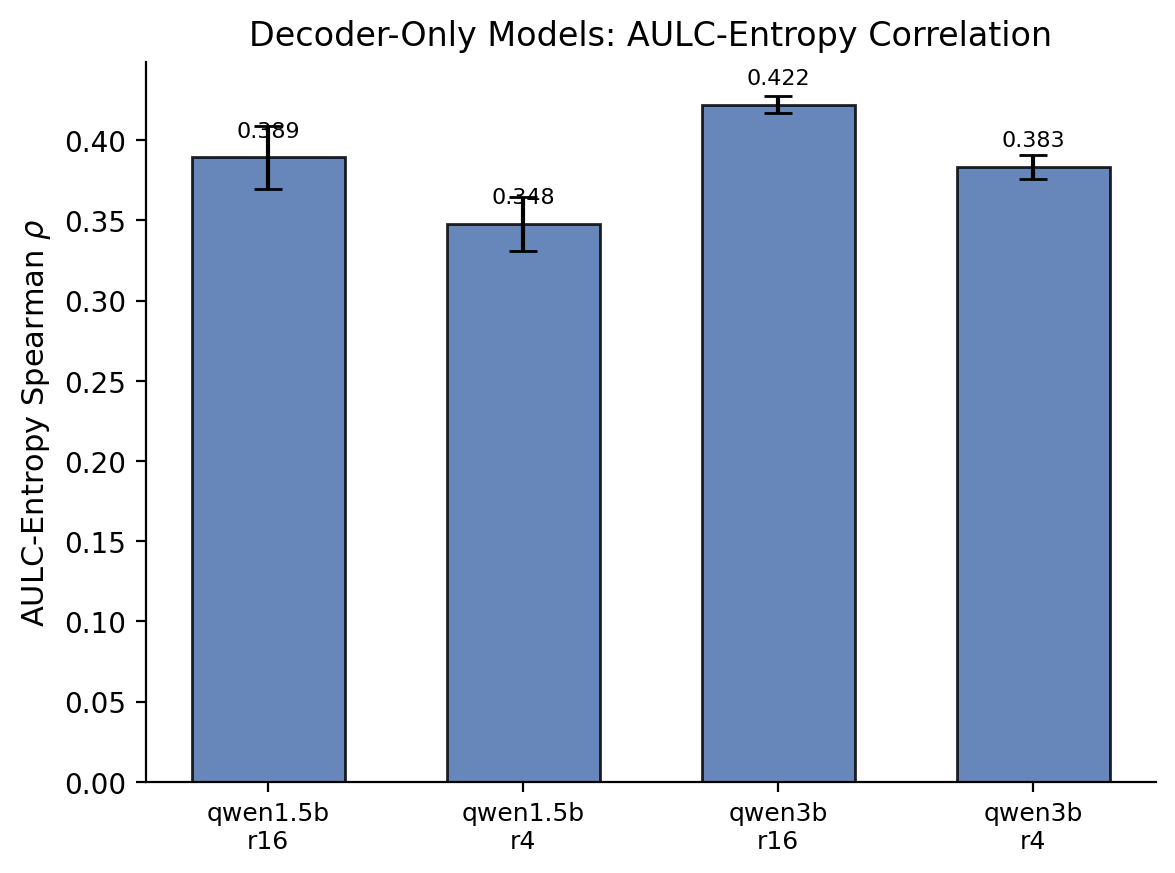}
\caption{Cross-architecture comparison of AULC--entropy
Spearman $\rho$ on SNLI.
Decoder-only models (Qwen2.5-1.5B, Qwen2.5-3B) show
stronger and more consistent correlations than the
encoder baseline (DeBERTa~v3).
Within each architecture family, higher LoRA rank
produces stronger correlations.
Error bars show $\pm$1 sample standard deviation across
seeds.}
\label{fig:decoder-summary}
\end{figure}

\section{Dataset Cartography Comparison}
\label{app:cartography}

\begin{figure}[H]
\centering
\includegraphics[width=\columnwidth]{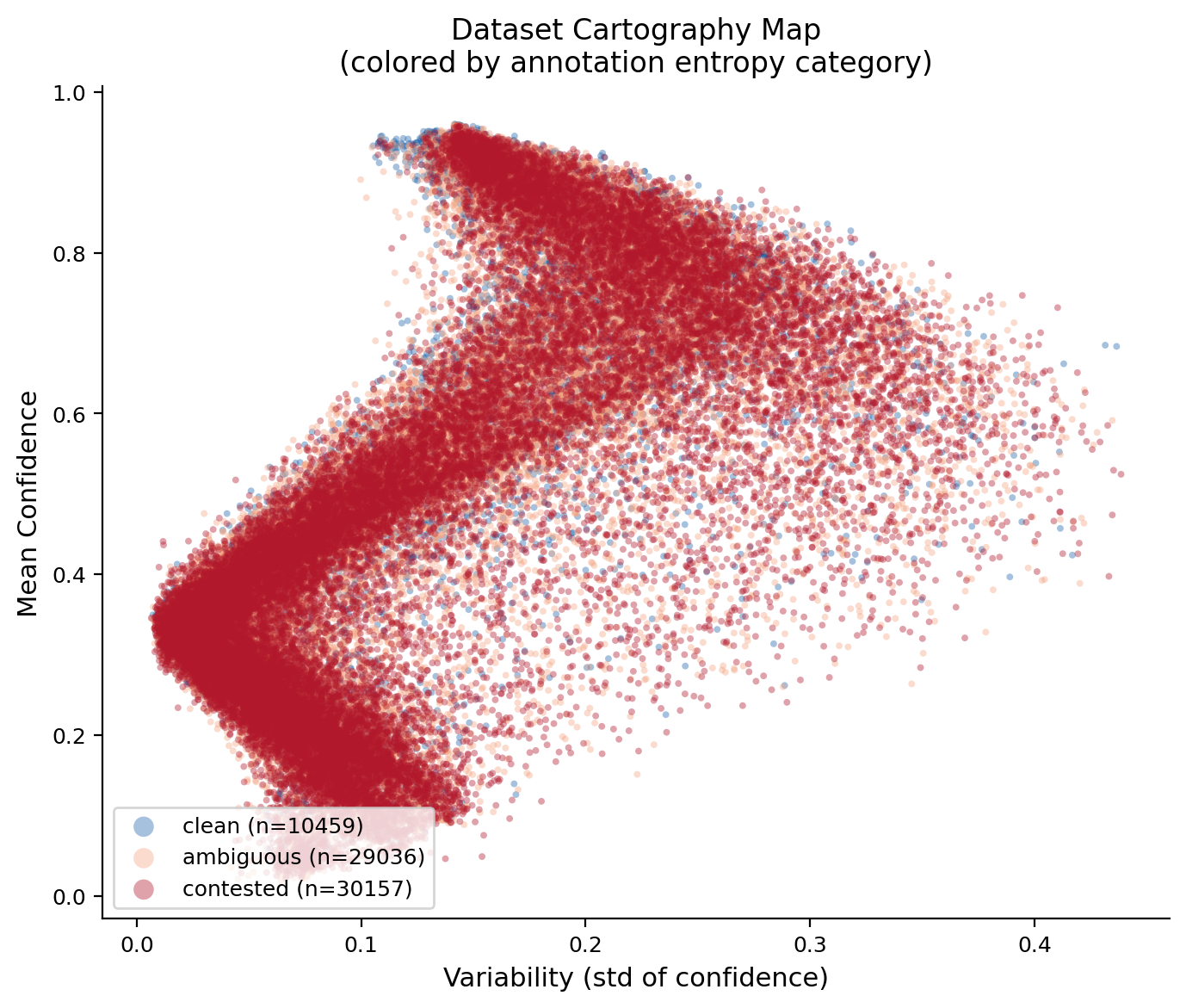}
\caption{Dataset Cartography map colored by annotation
entropy category (pooled across all available runs).
Clean examples (blue) cluster at high confidence, while
contested examples (red) are spread across the confidence--variability space.
The limited overlap between entropy-based and
cartography-based categories reflects their different
grounding: external human agreement vs.\ model-internal
training statistics.}
\label{fig:cartography-scatter}
\end{figure}

Figure~\ref{fig:cartography-scatter} plots the Dataset
Cartography map (confidence vs.\ variability) colored by
annotation entropy category, pooled across all available
tracker runs (54 main-matrix runs plus additional
rank-sweep configurations).
Clean examples cluster at high confidence / low variability,
while contested examples spread across the cartography space,
confirming that entropy categories and cartography regions
capture related but distinct structure.
The category overlap is limited: only 24\% of clean examples
fall in cartography's ``easy-to-learn'' region, and only 33\%
of contested examples fall in ``hard-to-learn,'' reflecting
the different bases of the two categorizations (external
annotation agreement vs.\ model-internal training statistics).

\section{Gradient Alignment Analysis}
\label{app:gradient-cosine}

\begin{figure}[H]
\centering
\includegraphics[width=\columnwidth]{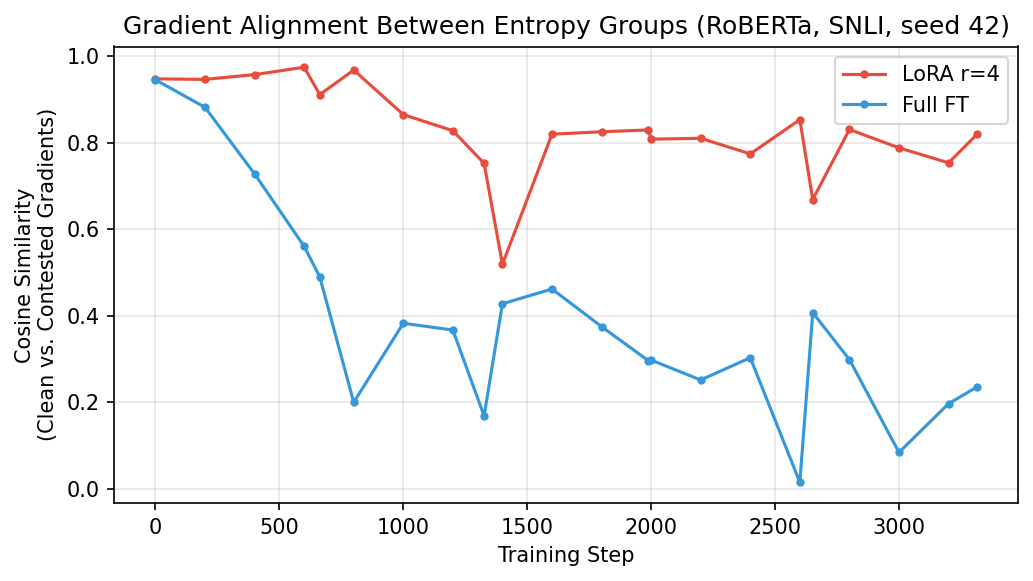}
\caption{Cosine similarity between aggregated gradients of
clean and contested examples over training
(RoBERTa, SNLI, seed 42).
Under LoRA $r{=}4$ (red), gradients remain well-aligned
(${\sim}0.80$), while under full fine-tuning (blue),
alignment drops to ${\sim}0.25$ by mid-training.
Un-learning under LoRA occurs despite high gradient alignment,
suggesting the bottleneck is subspace capacity rather than
gradient direction competition.}
\label{fig:gradient-cosine}
\end{figure}

Figure~\ref{fig:gradient-cosine} plots the cosine similarity
between aggregated gradients from clean and contested examples
over the course of training, for both LoRA $r{=}4$ and full
fine-tuning (RoBERTa, SNLI, seed 42).

Under LoRA $r{=}4$, gradients from the two entropy groups
remain well-aligned throughout training (cosine similarity
fluctuates between ${\sim}0.67$ and ${\sim}0.97$, with a
terminal value of ${\sim}0.82$).
Under full fine-tuning, alignment drops sharply from
${\sim}0.95$ at initialization to ${\sim}0.20$--$0.40$ after
the first ${\sim}800$ steps, reflecting the model's capacity
to learn group-specific update directions.

This pattern is counterintuitive: LoRA exhibits un-learning
\emph{despite} high gradient alignment, while full fine-tuning
resolves both groups \emph{despite} low alignment.
One interpretation is that the low-rank bottleneck forces
updates into a shared subspace where the net effect, though
directionally similar for both groups, disproportionately
benefits clean examples.
Full fine-tuning's larger parameter space can accommodate
divergent gradient directions, allowing it to improve on both
groups simultaneously.

\section{Calibration Analysis}
\label{app:calibration}

\begin{figure}[H]
\centering
\includegraphics[width=\columnwidth]{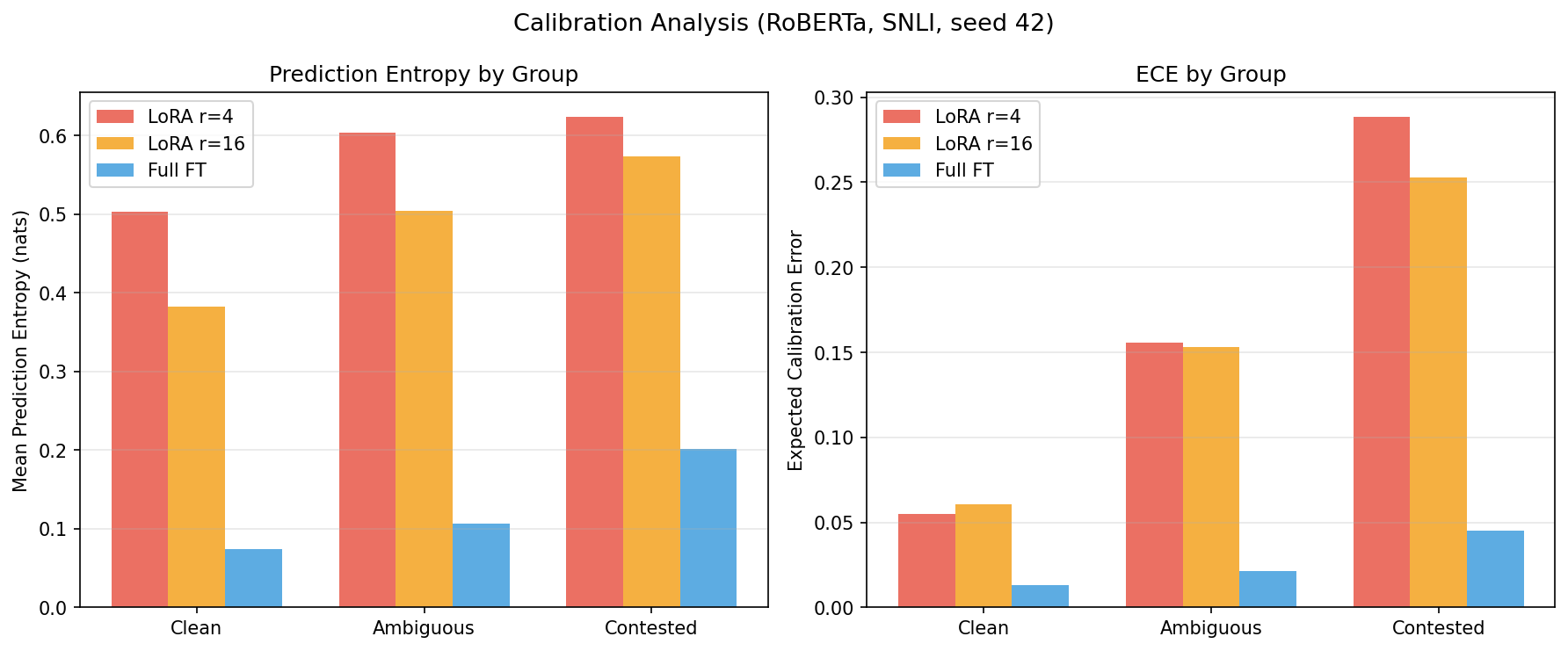}
\caption{Prediction entropy (left) and expected calibration
error (right) by entropy category
(RoBERTa, SNLI, seed 42).
Full fine-tuning produces far more peaked predictions
(lower entropy) than LoRA yet does not exhibit un-learning,
ruling out calibration as the primary explanation.
LoRA shows disproportionately high ECE on contested examples
relative to clean, reflecting the rank constraint's selective
impact on high-entropy items.}
\label{fig:calibration}
\end{figure}

Figure~\ref{fig:calibration} compares prediction entropy and
expected calibration error (ECE) across entropy categories for
LoRA $r{=}4$, LoRA $r{=}16$, and full fine-tuning (RoBERTa,
SNLI, seed 42).

Full fine-tuning produces substantially more peaked predictions
than either LoRA configuration: mean prediction entropy is
$0.12$ nats (full FT) vs.\ $0.49$ (LoRA $r{=}16$) and $0.58$
(LoRA $r{=}4$), and mean maximum confidence is $0.96$ (full FT)
vs.\ $0.80$ (LoRA $r{=}16$) and $0.76$ (LoRA $r{=}4$).
Full fine-tuning also achieves much lower ECE
($0.018$ vs.\ $0.149$ and $0.152$).

Critically, despite its far more peaked predictions, full
fine-tuning does \emph{not} exhibit un-learning
(Table~\ref{tab:unlearning}).
This directly rules out the calibration hypothesis, if
increasingly peaked predictions were the primary cause of
rising loss on contested examples, full fine-tuning should
show the strongest un-learning, not the weakest.
The ECE gradient across entropy categories is also informative:
under LoRA $r{=}4$, contested examples have ECE of $0.289$
vs.\ $0.055$ for clean (a $5.3{\times}$ gap), while under full
fine-tuning the gap is much smaller ($0.045$ vs.\ $0.013$,
$3.5{\times}$), confirming that LoRA's capacity constraint
produces systematically worse calibration specifically on
high-entropy examples.

\end{document}